\documentclass[a4paper]{article}

\usepackage[utf8]{inputenc}
\usepackage[T1]{fontenc}
\usepackage{lmodern}
\usepackage{mathdots}
\usepackage{xcolor,colortbl}
\usepackage{amsmath,amsthm,amsfonts} 
\usepackage{mathtools}
\usepackage{epsfig,amssymb,amsfonts,amsmath}
\usepackage{mathtools}
\usepackage{graphicx}
\usepackage{caption}
\usepackage{subcaption,todo}
\usepackage[a4paper,includeheadfoot,margin=2.54cm]{geometry}

\DeclareMathOperator*{\argmin}{arg\,min}

\long\def\ignora#1{\vskip0pt}

\usepackage{algorithmic}
\usepackage{algorithm}
\newcommand{\rank}[1]{{\rm rank}(#1)}
\newenvironment{code1}{%
                           \mathcode`\:="603A  
                           
                           \par
                           \upshape
                           \begin{list} 
                              {} {\leftmargin = 0.0cm}
                           \item[]
                           \begin{tabbing}
                              \hspace*{.3in} \= \hspace*{.3in} \=
                              \hspace*{.3in} \= \hspace*{.3in} \=

\hspace*{.3in} \= \hspace*{.3in} \= \kill
   }{\end{tabbing}\end{list}}

\theoremstyle{definition}

\definecolor{vlightgray}{gray}{0.95}
\title{ Adaptive Nonnegative Matrix Factorization and Measure Comparisons for Recommender Systems 
\author{Gianna M. Del Corso\thanks{Dipartimento di Informatica, Universit\`a di Pisa. \texttt{gianna.delcorso@unipi.it}, \texttt{francesco.romani@unipi.it}. This research  was partially supported by GNCS projects ``Metodi numerici avanzati per
		equazioni e funzioni di matrici
		con struttura'' and by University of Pisa under the grant PRA-2017-05.}  \and Francesco Romani\footnotemark[1]}}

\begin{document}
	
\maketitle
	
\begin{abstract}
The Nonnegative Matrix Factorization (NMF) of the rating matrix has shown to be an effective method to tackle the recommendation problem.  In this paper we propose new methods based on the NMF of the rating matrix and we compare them with some classical algorithms such as the SVD and the regularized and unregularized non-negative matrix factorization approach. In particular a new algorithm is obtained changing adaptively the function to be minimized at each step, realizing a sort of dynamic prior strategy. Another algorithm is obtained modifying the function to be minimized in the NMF formulation by enforcing the reconstruction of the unknown ratings toward a prior term.  We then combine different methods obtaining two mixed strategies which turn out to be very effective in the reconstruction of missing observations. We perform a thoughtful comparison of different methods on the basis of several evaluation measures. We consider in particular rating, classification and ranking measures showing that the  algorithm obtaining  the best  score for a given measure is in general the best also when different measures are considered, lowering the interest in designing specific evaluation measures. The algorithms have been tested on different datasets, in particular  the 1M, and 10M MovieLens datasets containing ratings on movies, the Jester dataset with ranting on jokes and Amazon Fine Foods dataset with ratings on foods. The comparison of the different algorithms, shows  the good performance of methods employing both an explicit and an implicit regularization scheme. Moreover we can get a boost by mixed  strategies combining a fast method with a more accurate one. 
\end{abstract}



\section{Introduction}

Consumers are literally submerged by large selections of products and
choices. Recommender Systems are tools designed to help retailers to find the most
appropriated products that meet the needs and tastes of their users. Content filtering and collaborative filtering are two alternative approaches to this interesting problem. The content-based filtering approaches try to recommend items that are similar to those that a user liked in the past~\cite{RSHB11}, whereas systems designed according to the collaborative filtering paradigm identify users whose preferences are similar to those of the given user and recommend items they have liked~\cite{BS97}.

Some of the most effective approaches to collaborative filtering  are the so called \emph{Latent Factor Models}~\cite{CKT10,DKM15,KBV09,SM08,KBV09,SKKR02b}. These models try to view the expressed ratings as characterized by a low number of factors inferred from the rating patterns to reduce the dimension of the space of the users and of the items. Mathematically, this corresponds in approximating the matrix containing the known ratings with a low-rank matrix and use the latter to infer the unknown ratings. Two popular numerical methods used in this context are the Singular Value Decomposition (SVD)~\cite{CKT10} and the Nonnegative Matrix Factorization (NMF)~\cite{KBV09}. The power of the SVD in  fields such as machine learning and data analysis is well known, and, as well explained in~\cite{Gi14}, the NMF shares many of these positive aspects and it is particularly well suited for features extraction and reconstruction of missing observations. Both these factorizations can be formulated as minimization problems that, in the case of NMF, is non-convex. The usual approach for computing the NMF is based on an alternating  non-linear least square scheme, where at each step we have to solve two convex subproblems. This approach has shown to lead to optimal solutions~\cite{KP08,Li07}. 

In this paper we propose new methods based on the Nonnegative Matrix Factorization (NMF) of the rating matrix and we compare them with some classical algorithms such as the SVD and the regularized and unregularized non-negative matrix factorization approach. In particular an algorithm is obtained modifying the function to be minimized in the NMF formulation by enforcing the reconstruction of the unknown ratings toward a prior term. Another algorithm 
is obtained changing adaptively the function to be minimized at each step, realizing a sort of dynamic prior strategy.  We then combine different methods obtaining two mixed strategies which turn out to be very effective in the reconstruction of missing observations.

Recommendation systems have been designed with different goals in mind. For example they have been used to predict missing ratings in order to make personalized recommendations~\cite{KBV09}, to generate a ranked list of items to submit to each user~\cite{BC12,CKT10}, or to classify items as interesting/not-interesting for each user. To capture these multiple goals, many different measures (metrics) for recommender systems have been introduced in the literature~\cite{CKT10,Fa06,NK12,STL11}. In this paper we compare old and new algorithms with respect to many different measures, considering rating, classification and ranking metrics for addressing all the different scenarios for which recommender system are designed. 

The extensive testing shows a number of interesting properties: for example the optimal number of latent factors is independent of the metric but is specific of the algorithm. Moreover, comparing the results for many problem instances we discovered that the different metrics correlate very well. For example, the Spearman correlation between ranking metrics is never lower than 94\%, 
raising some doubts about the need of so many different measures, at least in general. 

Our experiments on four different datasets show that the new methods proposed in this paper outperforms the classical schemes. In particular the method combining a static prior and regularized method makes consistently better prediction with respect to most of the metrics. We also observed, that, in order to make good predictions, it is sufficient to consider a relatively small number of latent factors.  



The paper is organized as follows: The  recommendation problem based on Matrix factorization is formally presented in Section~\ref{sec:MF}, while the algorithms proposed and analyzed in the paper are introduced in Section~\ref{sec:alg}. Section~\ref{sec:evmet} describes the different evaluation metrics considered in this paper, dividing them into rating, classification and ranking metrics.  Section~\ref{sec:exp} contains  the experimental results, addressing both the question of comparison of the measures as well as of the performance of the different methods. In Section~\ref{sec:concl} we draw some conclusions.

\subsection{Related works}

Numerical techniques have proven to be  very useful to design effective algorithms in many areas related to  information searching, ranking and retrieval~\cite{BDR10,DR16,FMRR13,Gl15,GG06,MMBB07,SHCGW17}. In particular, the idea of factorizing a matrix  to linearly reduce the dimensionality of the problem is a well known technique widely used in image processing~\cite{LS99}, text mining~\cite{PSBP}, classifications~\cite{BGG09} but also in design of effective recommender systems~\cite{CKT10,DKM15,KBV09,SM08}.  The literature on this subject is wide, we refer to~\cite{BGM15} and to the references therein for a survey on the matrix factorization models in collaborative filtering. 

In~\cite{CKT10}, the authors compare collaborative filtering algorithms and propose both neighborhood models and Latent Factor models algorithms based on the Singular Value Decomposition (SVD). We included the latter algorithm in our experiments denoting it PSVD.  In~\cite{Ga18} the authors proved that it is possible
to reformulate the PSVD algorithm as the computation of an eigendecomposition of a cosine similarity
matrix. A whole family of methods, which the authors called EigenRec, can be obtained choosing different
similarity measures such as the Pearson-Correlation and the Jaccard Similarity measures, and different scaling functions.  An efficient method, based on the Lanczos algorithm is proposed to build the latent space.
The SVD approach is considered also in~\cite{SKKR00c}, but in that case the missing evaluations are filled using the average  ratings for a user or the average ratings for an item instead of the zero value. Despite the filled rating matrix is dense, one can take advantage from the special structure of the additional entries and used a suitably modified sparse SVD routine to factorize the rating matrix. The authors found that a small number of latent parameters (14 for the MovieLens dataset of 100K ratings) is the most suitable respect to the MAE error measure (defined in Section~\ref{sec:evmet}). Despite our testing methodology is slightly different we obtained similar results on the number of latent factors. 

An different approach is the one pursued in~\cite{DKM15}, where the authors assume that it is more likely for an unknown item to be weakly rated because a user is generally interested in a limited list of items compared with the total number of existing items. For capturing this aspect, the authors propose to incorporate in the optimization problem a prior term which drives the unknown ratings toward a prior estimate $\alpha$. We include this model in our experimentation (see model~\eqref{nmf:formregprior}), but we noted that forcing the reconstruction of missing values toward the values zero is not satisfactory and better results can be obtained with a different value of $\alpha$.

Ning and Karypis~\cite{NK12} propose a sparse Linear Model algorithm for the top-$N$ recommendation problem (see Section~\ref{sec:evmet}). The recommendation scores are computed learning a positive weighting matrix $W$ and solving regularized least squares problem. 
In~\cite{Pan08} and~\cite{Hu08} Weighted Regularized Matrix Factorization (WRMF) methods are formulated as a Least-Squares problem. The weighting matrix is used to differentiate the contributions from observed activities and unobserved ones. In~\cite{ZLCLL17}  the authors formulate the optimization problem in terms of the $R_1$-norm instead of  the usual Frobenius or $L_2$ norm. Their interest is in designing a NMF-based method robust to malicious attacks where users inject fake profiles to manipulate the recommender system. Liu and Wu~\cite{LiWu16} propose a latent factor model transforming the recommendation problem into a nearest neighbor search problem. To this end, users and items are projected into the latent space, and similarities between items and users are then computed to provide recommendations. It was not possible a direct comparison of all the above approaches with our algorithms because of different testing methodologies and the use of different error measures.


\section{Matrix factorizations for recommender systems}	\label{sec:MF}

Formally, we use the following notation.
Let ${\cal U}=\{u_1, u_2, \ldots, u_m\}$ denote a set of users, ${\cal I}=\{ i_1, \ldots, i_n\}$ denote a set of items, and ${\cal V}\subseteq [v_{\min},  v_{\max}]$ denote the set of possible votes that a user can assign to an item. Define ${\cal V}_0={\cal V} \cup \{?\}$ the set of possible votes plus the value ? corresponding to the undefined or missing evaluation.
Let $A\in {\cal V}_0^{m\times n}$ be the {\it Utility Matrix} (also called rating matrix), and let $\Omega \subseteq {\cal U}\times {\cal I}$, $\Omega= \{(u, i)|\, a_{ui}\in {\cal V}\}$. Each element $(u, i)\in \Omega$  represents that  user $u$ has evaluated  item $i$, the corresponding vote (or rating) is stored in entry $a_{ui}$ of $A$. Let us denote by $\bar{\Omega}$, 
the complement of $\Omega$, the set containing  missing evaluations, i.e.  $\bar{\Omega}=\{(u, i)| \, a_{ui}=?\}$.

\emph{Latent Factor Models}~\cite{DKM15,KBV09,SM08} try to view the expressed
ratings as characterized by a low number of factors inferred from the rating
patterns to reduce the dimension of the space of the users and of the items. Low-rank matrix factorization of the rating matrix is a common technique for discovering the  latent factors and represent items and users in terms of a few vectors. The available data are used to compute the latent factors which are later employed  to predict ratings for items not yet rated by a user. This approach for predicting unknown ratings relies on the fact that a model accurately predicting observed ratings generalizes well to unknown ratings. In the following we propose some common approaches that will be the starting tool to present old and new  algorithms in Section~\ref{sec:alg}. 

Given the utility matrix $A$, and the set of expressed ratings $\Omega\subseteq {\cal U}\times {\cal I}$, we denote by $A_\Omega$ the matrix obtained from $A$ replacing~? with~0. More in general, given a set $X\subseteq {\cal U}\times {\cal I}$ and a matrix $M$, we  use the notation $M_X$ to represent the matrix obtained applying to $M$ a projection operator that only retains the entries lying  in the set $X$, i.e.
\begin{equation}\label{eq:proiettore}
M_X=\left\{\begin{array}{ll}
m_{ij} & \mbox{if } (i, j)\in X\\
0 &  \mbox{if } (i, j)\not \in X.
\end{array}\right.
\end{equation}
In the following we will denote by ${\bf e}_i$ the $i$-th column vector of the canonical basis, that is the vector with all components equal to zero except for the $i$-th that is equal to 1. \ignora{We will similarly denote by ${\bf u}$ the vector of ${\mathbb R}^n$ with all the entries equal to one. }

\subsection{The SVD approach}
When latent factor models were first proposed the SVD approach was the natural choice~\cite{KBV09}. In this case we look for a matrix $B$ of rank at most $k$ such that
\begin{equation}
\label{svdp}
\min_{\rank{B}\le k}{\|A_\Omega -B\|_F}.
\end{equation}
That minimum is achieved for the  SVD truncated at the $k$-th term, $B=P_k\Sigma_k Q_k^T$, being  $A_{\Omega}$, that is $A_{\Omega}=P\Sigma Q^T$.
Since the  columns of $P_k$ and $Q_k$ are orthogonal vectors, in general matrix $B$   will have some negative  entries that should be somehow interpreted when the set ${\cal V}$ does not allow negative ratings. For this reason	
many authors argue that SVD approach is not adequate for this problem.

An idea for partially avoiding this phenomenon, when working with nonnegative ratings, is to shift each entry of matrix $A_\Omega$ with a fixed $\gamma>0$, and then apply SVD. Denoting by ${\bf u}_m$ and ${\bf u}_n$ the vectors with entries equal to one and length equal to $m$, the number of users, and  $n$, the number of items,  we get
\begin{equation}\label{svds}
P_k\Sigma_kQ_k^T=\argmin_{\rank{C}\le k}\|A_\Omega-\gamma \,({\bf u}_m{\bf u}_n^T)_{\Omega} -C\|_F 
\end{equation}
where $P\Sigma Q$ is the SVD decomposition of $A_\Omega-\gamma \,({\bf u}_m{\bf u}_n^T)_{\Omega}$. 
We have then to shift back the values of matrix $P_k\Sigma_kQ_k^T$ adding the value $\gamma$ to each entry, that is set $B=P_k\Sigma_kQ_k^T+\gamma \,{\bf u}_m{\bf u}_n^T$.
In this case we have to choose both a suitable value for $k$ and the parameter $\gamma$.
Note that for moderate values of $k$ a sparse SVD can be carried out efficiently using  packages such as SVDPACK~\cite{Be92} or PROPACK~\cite{La01}.

\subsection{The Nonnegative matrix factorization approach}

When we are interested in a rank-$k$ approximation with non negative entries it is more convenient 
to consider  the Nonnegative Matrix Factorization (NMF). In the case the set ${\cal V}$ allows negative votes, we can simply shift the values to have an utility matrix containing only nonnegative entries. The Nonnegative Matrix Factorization problem
 can be mathematically formulated as follows~\cite{KP11}. Given a matrix $M\in \mathbb{R}^{m\times n}$ in which each element is nonnegative, i.e. $, m_{ij}\ge 0$, and an integer $k < \min \{m, n\}$, NMF aims to find two factors $W \in \mathbb{R}^{m\times k}$ and $H \in \mathbb{R}^{n\times k}$ with nonnegative elements such that  $WH^T$ is the closest matrix to $M$ with  respect to a suitable norm.
A common choice is  to consider the Frobenius norm, and the problem can be formulated as an optimization problem  where  $W$ and $H$  are found 
solving the  non-convex optimization problem 
\begin{equation}\label{eq:nmf}
\min_{H\geq 0, W\geq 0} F(H, W)=\frac{1}{2}\|M-WH^T\|_F^2.
\end{equation}
 Since problem~\eqref{eq:nmf} is non-convex we can reasonably expect to find only a local minimum. Many algorithms have been devised to solve this problem inside an Alternating Nonnegative Least Square (ANLS) framework, where the   non-convex minimization problem is formulated as a two coordinate descent problem~\cite{KP08}. 
Given an integer $k \ll n$ one of the factors, say $W$ is initialized with non-negative entries, and then an alternating constrained least square scheme is iterated until certain convergence criteria are met.

The general ANLS algorithm can be described as follows
\\
\medskip
\centerline{
	\framebox{\parbox{8.0cm}{
			\begin{code1}
				{\bf Procedure} {\bf ANLS }\\
				{\bf Input}:\quad$M$, $W_{0}$ \\
				\ {i:=0};\\
				\ {\bf repeat} \\
				\qquad $H_{i+1}:=\argmin_{H\ge 0} {\cal F }(X, W_i);$\\
				\qquad $W_{i+1}:=\argmin_{W\ge 0} {\cal F }( H_{i+1},X);$\\
				\qquad $i:=i+1$;\\
				\ {\bf until} stopping condition\\
				\\
				{\bf Output}:\\
				\quad $W_{i}$, $H_{i}$
			\end{code1}
}}}
where ${\cal F}$ is the function to be minimized that could be~\eqref{eq:nmf}, or another one containing regularizing or prior terms as we will see.

It can be proved that every limit point generated from the ANLS framework is a stationary point for the non-convex original problem~\cite{KP11}.
To solve the least square problems inside the ANLS framework we can use one of the many methods developed such as the Active-set method~\cite{KP08b}, the projected gradient method~\cite{Li07},  or the projected quasi-Newton method~\cite{KSD07}. In this paper we use  the greedy coordinate descent method~\cite{HD11} which we describe in detail in Section~\ref{sec:gcd}. This method is particularly well suited when the matrix to be factorized is sparse as in the case of recommender systems because we can take advantage from the sparse structure to implement it  in a convenient way.

In out setting, if we apply the NMF directly to $A_\Omega$ we force the reconstruction of the missing evaluations towards zero. Instead, since we are interested in a NMF that closely approximate the expressed ratings and returns a estimate of the missing ratings we formulate the optimization problem as follows
\begin{equation}\label{nmf:form}
 \min_{W\ge 0, H\ge 0} G(H, W)= \frac{1}{2} \| A_\Omega -(WH^T)_\Omega\|_F^2.
\end{equation}

Usually regularization terms are added to avoid overfitting of data. Using the 1-norm, which, with a slight abuse of notation we define as  $\|M\|_1=\|\mbox{vec}(M)\|_1=\sum_{ij} |m_{ij}|$, the model  becomes
\begin{equation}\label{nmf:formreg}
 \min_{W\ge 0, H\ge 0}L(H, W)= \frac{1}{2} \| A_\Omega-(WH^T)_\Omega\|_F^2+\lambda(\|H\|_1+ \|W\|_1).
\end{equation}
In general other norms can be used for the regularization parameter, but the advantage of working with the  1-norm is that it favors sparsity and it is very easy to implement.

Another kind of regularization can be obtained looking for terms $W$ and $H$ such that they agree with a prior term $\alpha$ on unknown ratings. The function to be minimized becomes
\begin{equation}\label{nmf:formregprior}
  \min_{W\ge 0, H\ge 0} P(H, W)=  \frac{1}{2}\, \| A_\Omega-(WH^T)_\Omega \|_F^2+ \mu\, \| ( \alpha\,{\bf u}_m {\bf u}_n^T- WH^T)_{\bar \Omega}\|_F^2. 
\end{equation}
In this model  the same scalar value of $\alpha$ is assumed to be a good prediction for all the unknown ratings.  

\subsection{Greedy coordinate descent algorithm} \label{sec:gcd}

The Greedy Coordinate Descent (GCD) algorithm was proposed in~\cite{HD11} to solve the least square problem inside an ANLS scheme. In the original paper the algorithm is presented for full matrices  but, as we will see in the following, it can exploit easily both sparsity and one-norm regularization, so that it turns out to be particularly suitable for recommender systems where the observed votes are only a small portion  respect to the entries of the matrix. Its speed features and guaranteed convergence makes it a reasonable choice inside a recommender system based on an ANLS method.

%

In our framework, denoting by ${\cal F}(H, W)$ the function to be minimized, that in our case is one among $G(H, W)$~\eqref{nmf:form} and $L(H,W)$~\eqref{nmf:formreg}, and choosing a pair $(u, i)$,   coordinate descent solves
the following one-variable subproblem to get $s$ such that
\begin{equation}\label{eq:fur}
\min_{s: w_{ur}+s >0} f_{ur}(s)^{W}={\cal F}(H, W+sE_{ur}),
\end{equation}
where $E_{ur}$ is an $m\times k$ matrix with  all entries zero except the $(u,r)$ element equal to one, i.e. $E_{ur}={\bf e}_u {\bf e}_r^T$.

In the case of the formulation given in~\eqref{nmf:form}, denoting by $R=R_{\Omega}$ the sparse residual matrix, i.e. $R=A_{\Omega}-(WH^T)_\Omega$, we have
$$ 
f_{ur}(s)^{W}=\frac{1}{2}s^2\sum_{j\in \Omega_u} h_{jr}^2 -s \sum_{j\in \Omega_u} r_{uj} h_{ur}+\frac{1}{2}\sum_{j\in\Omega_u}r_{uj}^2,
$$
where $\Omega_u=\left\{ j |\, a_{uj}\in {\cal V}\right\}$, is the set of  items voted by user $u$. Since $f_{ur}(s)^{W}$ is a degree two polynomial in $s$, the minimum is achieved for $$\bar s=\left(\sum_{j\in \Omega_u} r_{uj} h_{ur}\right)/\left(\sum_{j\in \Omega_u} h_{jr}^2\right).$$ Hence, because of the nonnegative constraint, we get
$$
s^*=\left\{ \begin{array}{ll} \bar{s} & \mbox{ if } w_{ur}+\bar s >0\\
-w_{ur} & \mbox{ if } w_{ur}+\bar s \le 0.
\end{array}\right.
$$
The gain in the objective function results
$$
{\cal F}(H, W)-{\cal F}(H, W+s^*E_{ur})=-\frac{(s^*)^2}{2}\sum_{j\in \Omega_u} h_{jr}^2 + s^*\sum_{j\in \Omega_u} r_{uj} h_{ur},
$$
and we can update the residual matrix as follows
$$
r_{uj}=r_{uj}-s^*h_{uj}, \quad  \mbox{for all } j\in \Omega_u. 
$$

The formulation for the GCD algorithm is slightly modified when regularization in the one-norm is introduced as in problem~\eqref{nmf:formreg}. In this case the function to be minimized is 
$$
L(H, W)=\frac{1}{2} \| R\|_F^2+ \lambda \|H\|_1+\lambda \|W\|_1.
$$
and the value of the parameter $s$ minimizing $f_{ur}(s)^{W}$ is given by $$\bar s=(\lambda+ \sum_{j\in \Omega_u} r_{uj} h_{ur})/(\sum_{j\in \Omega_u} h_{jr}^2).$$

As we will see in Section~\ref{sec:alg:cost}, using the ideas presented in~\cite{DKM15} for dealing with the priors while maintaining the sparsity of the matrix, we can reformulate the GCD algorithm also for $P(H, W)$ in~\eqref{nmf:formregprior}.

Paper~\cite{HD11} describes how to embed this inner step into an iterative method  that reduces the objective function of a prescribed quantity. Note that  the  GCD algorithm  does not solve exactly the convex constrained minimization problems but it stops when the error has reduced of a factor $\varepsilon$. If $\varepsilon$ is not too small, this procedure turns out, once embedded into an ANLS scheme, to be much faster than other solutions such as the active-set method which at each step solves completely the convex problem. 
For further reference we denote by  ${\mbox{GCD}_1({\cal F}(X, W), \varepsilon)}$ the algorithm that given the function  ${\cal F}(X, W)$ and the tolerance $\varepsilon$, returns the matrix  which is an approximation depending on the parameter  $\varepsilon$ of $\argmin_{X\ge 0}{\cal F}(X, W)$. Similarly we denote by ${\mbox{GCD}_2({\cal F}(H, X), \varepsilon)}$ the algorithm returning a nonnegative approximation of  $\argmin_{X\ge 0}{\cal F}(H, X)$. A detailed description of these algorithms is given by Algorithm 1 (steps 1-8) in~\cite{HD11}, where the authors provide also a detailed analysis of the cost.

\section{The tested Algorithms}\label{sec:alg}

In this Section we introduce the algorithms we implemented and tested respect to the evaluation metrics we are going to describe in Section~\ref{sec:evmet}. For each algorithm we mention the parameters involved and how regularization is realized.

\begin{itemize} 
	\item {\bf PSVD} is the pure SVD approach of~\eqref{svdp}. Regularization is achieved taking a moderate value of $k$. 

	\item {\bf SSVD($\gamma$)} Is the shifted SVD approach of~\eqref{svds}. The shift is considered to overcome the problem of the negative entries in the reconstructed matrix.  Again regularization is obtained with a fairly low value of $k$. An adequate value of $\gamma$ should be chosen so that the matrix $B$ contains mostly  nonnegative entries. Taking $\gamma=0$ we obtain the {PSVD} approach above (see~\eqref{svdp}). Regularization is automatically realized keeping a low the value of the rank $k$. 
	\item{\bf NMF} Is the un-regularized Nonnegative matrix factorization approach described by~\eqref{nmf:form}. Since we employ the Greedy Coordinate Descent algorithm described in Section~\ref{sec:gcd} the tolerance $\varepsilon$ becomes an input parameter as well as the number of iterations $it$ of the ANLS scheme. The algorithm becomes
	
	\medskip
\centerline{
	\framebox{\parbox{8.0cm}{
			\begin{code1}
				{\bf Procedure} {\bf NMF }\\
				{\bf Input}:\quad$A$, $k, it, \varepsilon$ \\
				\\
				\ $W_0:=0_{m\times k}$;\\
				\ {\bf for} $i=1:it$ \\
				\qquad $H_{i}:={\mbox{GCD}_1({G}(X, W_{i-1}), \varepsilon)};$\\
				\qquad $W_{i}:={\mbox{GCD}_2({G}(H_{i}, X), \varepsilon)};$\\
				\ {\bf endfor} \\
				\\
				{\bf Output}:\\
				\quad $W_{it}$, $H_{it}.$
			\end{code1}
}}}
\smallskip

\item {\bf RNMF($\lambda$)} Is the Regularized Nonnegative matrix factorization approach described by~\eqref{nmf:formreg}. Beside the choice of $k$ we have to choose  $\lambda$, $\varepsilon$ and {\em it }. {RNMF} coincides with {NMF} when $\lambda=0$, and the algorithmic formulation is the same as procedure {NMF} calling the GCD algorithms using as  parameter function  $L$ rather than $G$ . 
%
%
\end{itemize}

In addition to the above known algorithms, in this paper we propose and test some new algorithms which use a prior term to improve the quality of the reconstruction. Within the framework of equation~\eqref{nmf:formregprior}, we propose  different strategies for selecting the parameter $\alpha$. 
\begin{itemize}
	\item {\bf PR($\alpha$)}  The algorithmic scheme is similar to that of Procedure {NMF}  but using function $P$ inside the GCD calls.  The prior term $\alpha$ acts also as a regularizing term which avoids the uncontrolled growth of the values of $W$ and $H$. If we set $\alpha=0$, we assign to all the unknown values the value zero, and the model forces the reconstruction of missing values toward this solution. 
This is the original proposal of~\cite{DKM15}, since they claim that, in general, items not rated will be likely to receive a weak rating. However, this does not apply when, for example, there are many items and a user is unlikely to visit all of them. 
For this reason, we also tested the algorithm with  $\alpha=(v_{\min}+v_{\max})/2$, which represent the neutral rating.
	\item {\bf PRD} (short for Dynamic with prior). 
	The idea is to use a scheme with an evolving prior factor. In particular at each iteration we assign to the unknown values  the estimate obtained by the previous iteration. Denoting by
\begin{eqnarray}\label{eq:fhpd}
	F_1(X,W, H)&=& \frac{1}{2}\, \| A_\Omega-(WX^T)_\Omega \|_F^2+ \frac{1}{2}\,\|(WH^T-WX^T)_{\bar\Omega}\|_F^2 \\
	F_2(H,X, W)&=& \frac{1}{2}\, \| A_\Omega-(XH^T)_\Omega \|_F^2+ \frac{1}{2}\,\|(WH^T-XH^T)_{\bar\Omega}\|_F^2 \nonumber 
	\end{eqnarray}
%
the formulation in the ANLS scheme becomes 	 
\\	

\smallskip
\centerline{
	\framebox{\parbox{8.0cm}{
			\begin{code1}
				{\bf Procedure} {\bf PRD }\\
				{\bf Input}:\quad$A, k, it, \varepsilon$ \\
				\\
				\ $W_0:=0_{m\times k}$;\\
				\ {\bf for} $i=1:it$ \\
				\qquad $H_i={\mbox{GCD}_1(F_1(X,W_{i-1}, H_{i-1}), \varepsilon)};$\\
				\qquad $W_i={\mbox{GCD}_2(F_2(H_i,X, W_{i-1}), \varepsilon)};$\\
				\ {\bf endfor} \\
				\\
				{\bf Output}:\\
				\quad $W_{it}$, $H_{it}$
			\end{code1}
}}}
\smallskip
%
This correspond to apply the standard NMF to an evolving matrix $B_i=A_\Omega+ (W_{i-1}H_{i-1}^T)_{\bar\Omega}$. Note that although $B_i$ is a full matrix, we can adapt the GCD 
 scheme~\eqref{eq:fur} to efficiently deal with the rank-$k$ term without building $B_i$.   
The evolving prior term, a moderate value of $k$ and a low number of iterations of the GCD scheme, contribute together to the regularization of the solution.  
	
\end{itemize}
Finally we tested also some mixed strategies,  setting  $\alpha_m=(v_{\min}+v_{\max})/2$, 
\begin{itemize}
	\item {\bf MIXR($h$)} where we start with $h$ steps of { PR($\alpha_m$)} and then we continue with {RNMF($\lambda$)}.  
	\item {\bf MIXD($h$)} where we start with $h$ steps of  { PR($\alpha_m$)} and then we continue with {PRD}.
\end{itemize}
Other mixed strategies, as well a full scheme  using both regularization and prior terms are possible, but the results obtained do not seem worth of reporting.  

\subsection{Considerations about the complexity} \label{sec:alg:cost}
To analyize the computational complexity of the different algorithms, let us denote by $s$ the number of nonzeros of the matrix $A_{\Omega}$, that is $s=| \Omega|$.  Then  the overall cost of {PSVD} and {SSVD}($\gamma$) is  $O(k\,T_{\mbox{mult}}+(m+n)k^2)$ floating point operations~\cite{HMT11}, where $T_{\mbox{mult}}$ denotes the cost of a matrix-vector multiplication and  hence in out setting  $T_{\mbox{mult}}=O(s)$.

The cost of each step of the algorithms based on nonnegative matrix factorization is given by $(T_{\mbox{GCD}_1}({\cal F}, \varepsilon)+T_{\mbox{GCD}_2}({\cal F}, \varepsilon))$, where ${\cal F}$ is either $G, L, P$ or the function defined by~\eqref{eq:fhpd}. From the analysis carried on in~\cite{HD11} we know that the cost of the GCD scheme  depends on   $s=|\Omega|$, on the number $t=t(\varepsilon)$ of subproblems~\eqref{eq:fur} we have to solve, as well as on the value of $k$. In our experiments,  $t$ turns  out to be increasing with $k$ and $m$ or $n$ depending on the fact we are updating $W$ or $H$, but always $t<mk^2$. For this reason (see for the details~\cite{HD11}),  in our case we get $T_{\mbox{GCD}_1}({\cal F}, \varepsilon)=O(mk^3)+ T_{\mbox{init}}$ and $T_{\mbox{GCD}_2}({\cal F}, \varepsilon)=O(nk^3)+T_{\mbox{init}}$, where $T_{\mbox{init}}=O(sk+k^3)$ is the time for the initialization of the matrices in the algorithm. Hence the asymptotic complexity of both  {NMF} and {RNMF}$(\lambda)$ is  $O\left( (n+m)k^3+sk\right)$ per iteration.

Under the reasonable assumptions $s>k^2$, the asymptotic cost per iteration  is the same also for procedures PR($\alpha$) and PRD.  The only difference could be the cost of the initialization phase of $\mbox{GCD}_i, i=1, 2$. However, as explained in~\cite{DKM15}, for the function $P$ in~\eqref{nmf:formregprior} we can exploit the rank-1 structure of the matrices involved to perform initialization in 
$O(sk+ (m+n)k)$ operations. The same reasoning can be applied for the {PRD} method, where the matrix involved is a rank-$k$ modification of a sparse matrix.

\section{Evaluation metrics} \label{sec:evmet}

Metrics for evaluating the performance of recommender systems can be classified in three main classes~\cite{STL11} depending on the final goal of the recommender system: rating metrics, classification metrics or ranking metrics. Since we can only measure the quality of an algorithm by comparing the algorithm estimates with actual data, in each experiment we partition the set of expressed ratings $\Omega$ in a training set ${\cal T}$ and test set ${\cal R}$ with ${\cal T}\cup{\cal R}=\Omega$ and ${\cal T}\cap{\cal R}=\emptyset$. Then, we apply the algorithms to matrix $A_{\cal T}$ producing a prediction matrix $B$ and evaluate the algorithm performance comparing $A_{\cal R}$ with $B_{\cal R}$. 

\subsection{Rating metrics}

Rating metrics~\cite{STL11} try to estimate how close the estimated ratings $B_{\cal R}$ are to the true user rating $A_{\cal R}$: they are used when it is important to predict the rating of all items. In this class we consider the {\it Mean Absolute Error} (MAE),
\begin{equation}\label{MAE}
\mbox{MAE}=\frac{1}{| {\cal R}|} \sum_{(u, i)\in {\cal R}} |a_{ui}-b_{ui}|,
\end{equation}
and the {\it Root Mean Square Error} (RMSE),
$$
\mbox{RMSE}=\sqrt{\frac{1}{| {\cal R}|} \|A_{\cal R}-B_{\cal R}\|^2_F}.
$$
Other popular rating metrics are the {\it Mean Square Error} (MSE), the {\it Normalized Mean Absolute Error} (NMAE), and the {\it Constrained Mean Absolute Error} (CMAE). 

\subsection{Classification metrics}
\begin{figure}
\begin{center}
\includegraphics[scale=0.7]{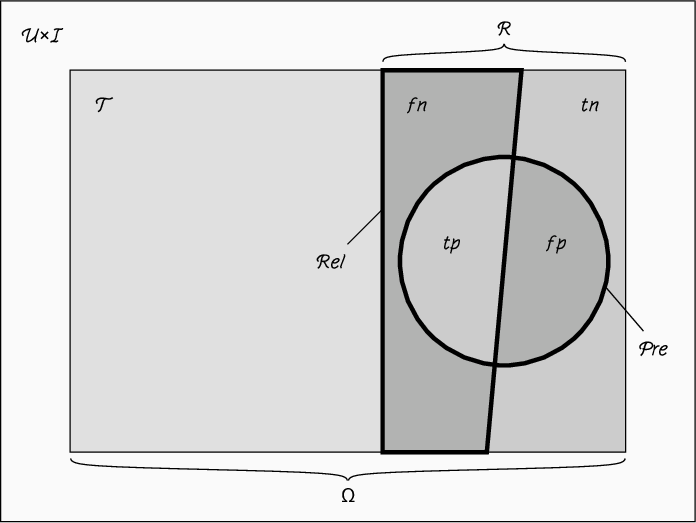}
\caption{Pictorial representation of the sets used for defining the different metrics. Note that    $\Omega={\cal R} \cup {\cal T}$. }\label{figtp}
\end{center}
\end{figure}

These metrics~\cite{STL11} are used when ratings are interpreted as a binary judgment (like/unlike). They are based on the comparison of $A_{\cal R}$ and $B_{\cal R}$ using two thresholds $\sigma_A$ and $\sigma_B$. These thresholds depend on the data, and on the maximum vote $v_{\max}$. For example, if votes are on a scale 1-5, a reasonable choice for $\sigma _A$ is 4. Depending on the algorithm used to retrieve missed votes, we can set $\sigma_B=\sigma_A$ but a different threshold can be used, for example if the range of values returned in $B$ is different. The principal measures can be defined on the basis of the two sets below:
\begin{itemize}
	\item {Relevant}:  {\em Rel} $ =\{(i, j)\in {\cal R}| \, a_{ij}\geq\sigma_A$\},
   \item {Predicted}: {\em Pre} $ =(i, j)\in {\cal R}| \, b_{ij}\geq\sigma_B$\}.
\end{itemize}
The perfect prediction is when the two classes coincides. Since perfect algorithms are unlikely, it is customary to define the following sets, pictorially represented in Figure~\ref{figtp} 
\begin{description}
	\item[{True positive:}] {\em tp} = {\em Rel\,}$\cap${\em Pre},
	\item[{False positive:}] {\em fp} = {\em Pre\,}$\cap(\overline{\mbox{\em Rel\,}}\cap{\cal R})$,
	\item[{True negative:}] {\em tn} = $(\overline{\mbox{\em Rel\,}}\cap{\cal R})\setminus${\em fp},
	\item[{False negative:}] {\em fn} = {\em Rel\,}$\setminus${\em tp},
\end{description}
and the following measures: 
\begin{description}
	\item[P]= $\displaystyle{\frac{|{\mbox{\em tp}}|}{|{\mbox{\em Pre}}|}}=\displaystyle{\frac{|{\mbox{\em tp}}|}{|{\mbox{\em tp}}|\, +\,|{\mbox{\em fp}}|}}, \quad$ {\em Precision}
	\item[R]= $\displaystyle{\frac{|{\mbox{\em tp}}|}{|{\mbox{\em Rel}}|}}=\displaystyle{\frac{|{\mbox{\em tp}|}}{|{\mbox{\em tp}|}\, +\,|{\mbox{\em fn}}|}}, \quad$ {\em Recall}
	\item[F]= $\displaystyle{\frac{|{\mbox{\em fp}}|}{|{({\cal R}\setminus\mbox{\em Rel})}|}}=\displaystyle{\frac{|{\mbox{\em fp}|}}{|{\mbox{\em fp}|}\, +|{\mbox{\em tn}}|}}, \quad$ {\em Fallout}
	\item[F1]= $\displaystyle{\frac{2\,{\bf P}\cdot {\bf R}}{{\bf P}+{\bf R}}}, \quad$ {\em F1-score}
	\item[A]= $\displaystyle{\frac{|{\mbox{\em tp}}|\, +\,|{\mbox{\em tn}}|}{|\cal R|}},\quad $ {\em Accuracy}.
\end{description}

\subsection{Ranking metrics}

Ranking (prediction) metrics measure how accurate  an algorithm is in predicting recommendations in the right preference order for the users. These measures are preferred when it is not important the exact values of the prediction but only their relative ranking.\cite{STL11} .


Let us define the set of relevant items for the user $u$ as  ${I_r}^{(u)}=\{i \in{\cal I}| (u, i)\in {\cal R},  \, a_{ui}\ge\sigma_A\}$. Note that it is possible for a given user,  to have that ${I_r}^{(u)}=\emptyset$. The set of users which have a non empty relevant item list, will be denoted as $U_r=\left\{ u\in {\cal U}|\,I_r^{(u)}\neq \emptyset \right\}$. In the following let ${\cal R}_u=\{i\in {\cal I}| (u, i)\in {\cal R}\}$, and  let  $\phi^{(u)}$ be the permutation of the indices in ${\cal R}_u$ induced by ordering the entries of the $u$-th row of $A_{\cal R}$ in a non increasing way, that is 
$$
a_{u\phi_{i-1}^{(u)}}\ge a_{u\phi_{i}^{(u)}}, i=2, \ldots, |{\cal R}_u|.
$$
The analogues sequence for matrix $B_{\cal R}$ is denoted by $\pi^{(u)}$, i.e.     
$$
b_{u\pi_{i-1}^{(u)}}\ge b_{u\pi_{i}^{(u)}}, i=2, \ldots, |{\cal R}_u|.
$$	
For a given user $u$, we define the precision for that user as the number of items actually relevant for that user, among the top $N$ items recommended to  $u$. 
We get
$$
\mbox{pr}_N^{(u)}=\{h\le N\, |\, \pi_h^{(u)}\in I_r^{(u)} \},
$$	
that is, $\mbox{pr}_N^{(u)}$ is the set of the relevant items for user $u$ in the top $N$ positions.
The {\it Average Precision} for user $u$ is then defined as
$$
 AP_u=\frac{1}{|I_r^{(u)}|} \sum_{i =1}^{| I_r^{(u)}|} \frac{|\mbox{pr}_i^{(u)}|}{i}
$$
and averaging over all the users with a non empty relevance list we get the {\it Mean Average Precision} (MAP)
$$
\mbox{MAP}=\frac{1}{|U_r|}\sum_{u\in  U_r} AP_u.
$$

This metric, as similar others we can define taking the geometric, harmonic or the quadratic mean instead of the arithmetic one, emphasizes true positives which appear at the top of the list. 

Accuracy can be also estimated by the ROC curve ({\it Receiver Operating Characteristic})~\cite{Fa06} which provides a graphical representation for the performance of a recommender system. The ROC curve plots for different set sizes, recall versus fallout. The ideal recommender system will go straight to a recall of 1 and  a fallout of 0 and then remain at that value until all the size of the set of  recommendations  equals the set of relevant items. A single measurement of the quality of recommendation is given by the AUC ({\it Area Under the Curve})~\cite{St10}: the better the recommender system, the higher is the AUC. 

Another measure which has received much attention is the {\em Normalized Discounted Cumulative Gain}~\cite{Li07} which
rewards methods for which the permutations $\pi$ agrees with $\phi$ for the top positions, while relevant items
ranked at low positions of the ranking contribute less to the final score than relevant items at top positions.


For some authors~\cite{CKT10,NK12} the goal of a recommender system is to find a few specific number of items which are supposed to be most interesting for a user. Some of the ranking measures above can be defined to account only for the first $N$ positions, so we get the P$@N$, R$@N$ and F$_1@N$ where the measure is computed on the first $N$ positions of $\phi^{(u)}$ and $\pi^{(u)}$. The NDCG$@N$ can be defined similarly.
%

\section{Numerical Experiments}\label{sec:exp}

We compared the algorithms and the metrics on four different datasets, whose characteristics are summarized in Table~\ref{tab:ds}. In particular we used the 1M, and 10M MovieLens datasets~\cite{Movie} containing evaluations on movies, the Jester dataset containing continuous ratings on 100 jokes~\cite{jester} and the Amazon Fine Foods dataset~\cite{finefoods} with reviews on foods. We removed from the Fine Foods dataset users with less than 5 evaluations and items which have been evaluated by only one user. The datasets have different characteristics, such as the ratio between users and items, which is high for Jester, moderate for the MovieLens datasets, and smaller than one for the Fine Foods dataset. Also the densities are very different, for example in Jester we have that more than half of the ratings are expressed, where the Fine Foods dataset has less than 4 expressed ratings over 10,000 entries in  the utility matrix. 

\begin{table}[!h]
	\begin{center}
		\begin{tabular}{|l|r|r|r|r|r|}
			\hline
			Dataset&\#Users&\#Items&\#Ratings&Density&Rating Scale\\ \hline
			Movielens 1M&6,040&3952&1,000,209&4.19 \% &1-5 discrete\\
			Movielens 10M& 71,567&65,133&10,000,054 &0.21\% &0.5-5 discrete\\
			Jester & 73,421&100&4,136,360&53.34\%&[-10, 10] continuous\\
			Fine Foods&11,985 &72,551 &316,010 &0.04\% & 1-5 discrete\\
			\hline
		\end{tabular}
		\caption{Different datasets used in the paper. We have datasets with different characteristics, for example with more users than items, or viceversa, or  with very different  densities. Moreover we have continuous or discrete ratings.}\label{tab:ds}
	\end{center}
\end{table}

As typically done in the literature, we adopt the $t$-fold cross validation methodology~\cite{GWHT14}[Chapter 5]. This approach consists in partitioning the ratings  into $t$ groups of approximately equal size. One group is used as test set and the remaining $t-1$ groups  are used to train the model. Each metric is then computed on the data in the test set. This procedure is repeated $t$ times, each time using a different group as test set. The final $t$-fold cross validation estimate is computed averaging the $t$ values obtained on the $t$ different test sets. Usually the value of $t=5$ is used meaning that the test set-training set ratio is  20\%-80\%. 
This  seems a good compromise between reducing the error due to bias and the error of the data. Using  a small value of $t$, we can overestimate the error because the training set contains not enough observations to predict the model. On the contrary, using a high value of $t$  the error due to the variance of the data becomes large,  since the $t$ values to be averaged are likely  less correlated due to the smaller  overlap between the training sets in each sample.  However, as we report in Section~\ref{sec:comp} for the datasets considered, we do not have significate changes in the behavior of the error measures with any $t$ between $3$ and $6$.

%

In addition to the algorithms described in Section~\ref{sec:alg}, we measured the performance of a RANDOM algorithm that assigns random evaluations to the pairs $(u,i)$ in the test set ${\cal R}$. The rationale of this experiments is that the scores produced by different measures have different distributions, so the comparison with a random algorithm is necessary to assess the quality of the proposed algorithms. For example, for the NDCG measure, it has been empirically observed and theoretically understood~\cite{Wang13} that NDCG converges to 1 almost surely as the size of the dataset increases even for random evaluations.

\subsection{Measure correlations and comparisons}\label{subsec:mc}

As we described in Section~\ref{sec:evmet} recommender systems can be designed with different purposes; for this reason different evaluation metrics have been proposed in the literature. A first contribution of our analysis is to compare the metrics reviewed in Section~\ref{sec:evmet} to see if they are somehow equivalent or if they highlight different features. 
\begin{table}[!h]
	\begin{center}{\footnotesize
			\begin{tabular}{|l| ll | ll | lllll|}
				\hline
				&{\em RMSE}&{\em MAE}& {\em F$_1$} &{\em A}& {\em MAP} & {\em AUC}& {\em NDCG} & {\em F$_1@10$} &{\em NDCG}@10  \\ \hline
				{\em RMSE}&\cellcolor{lightgray}100&\cellcolor{lightgray}90.2&75.3&83.8 &91.7&92.9&88.&91.&89.6 \\
				{\em MAE}&\cellcolor{lightgray}90.2&\cellcolor{lightgray}100&93.4&98.1 &86.5&88.6 &80.2&88.6 &83.8\\ \hline
				{\em F$_1$}&75.3&93.4&\cellcolor{lightgray}100&\cellcolor{lightgray}96.5&69.6 &72.7&61.8&73.1 &66.4\\
				{\em A}&83.8 &98.1 &\cellcolor{lightgray}96.5&\cellcolor{lightgray}100&79.9&82.2&72.4&84.3 &77.5\\ \hline
				{\em MAP}&91.7&86.5&69.6 &79.9&\cellcolor{lightgray}100&\cellcolor{lightgray}99.3 &\cellcolor{lightgray}98.2&\cellcolor{lightgray}97.9&\cellcolor{lightgray}99.1\\ 
				{\em AUC}&92.9&88.6 &72.7&82.2&\cellcolor{lightgray}99.3 &\cellcolor{lightgray}100&\cellcolor{lightgray}96.8 &\cellcolor{lightgray}97.6 &\cellcolor{lightgray}97.7\\
			$\mbox{NDCG}$ &88.&80.2&61.8&72.4&\cellcolor{lightgray}98.2&\cellcolor{lightgray}96.8 &\cellcolor{lightgray}100&\cellcolor{lightgray}94.5&\cellcolor{lightgray}98.9\\
				{\em F$_1@10$} &91.&88.6 &73.1 &84.3 &\cellcolor{lightgray}97.9&\cellcolor{lightgray}97.6 &\cellcolor{lightgray}94.5&\cellcolor{lightgray}100&\cellcolor{lightgray}97.7\\
				$\mbox{NDCG@10}$ &89.6 &83.8 &66.4&77.5&\cellcolor{lightgray}99.1 &\cellcolor{lightgray}97.7&\cellcolor{lightgray}98.9&\cellcolor{lightgray}97.7&\cellcolor{lightgray}100\\ \hline
			\end{tabular}
			\caption{Spearman ranks correlations for the different measures. We see that measures of the same kind correlate very well, but also rating metrics and ranking metrics are well correlated, while classification metrics such as  {\em F$_1$} and {\em A} have a good correlation with rating metrics but not so good with ranking metrics. \label{tab:speraman}}}
	\end{center}
\end{table} 

To test the correlation of the different metrics we performed an experiment on the MovieLens 1M dataset.  We selected 500 runs of the different algorithms proposed in Section~\ref{sec:alg} varying the methods and the parameters involved such as the rank $k$, the number of iterations inside the ANLS scheme, and the value of the regularization or prior parameters. For each of the metrics MAE, RMSE, F$_1$, A, MAP, AUC, NDCG, F$_1@10$ and NDCG$@10$, we obtain a vector of length 500, with the values of the metric for each run. We then computed the Spearman rank correlation~\cite{Sp04} of each metric against the others and we report the results in Table~\ref{tab:speraman}.

We see that the correlations between ranking metrics is higher than 94\%, suggesting that these metrics behaves similarly. Also classification and rating metrics are well correlated inside their respective class. Moreover, from Table~\ref{tab:speraman} we note that rating and ranking metrics are not so badly correlated as claimed in~\cite{CKT10}, for example the correlation between RMSE and all the ranking measures is never lower than 88\%. The correlation between MAE and A is as high as 98$\%$.

These considerations suggest us to introduce some cumulative measures, one for each kind of evaluation metrics, obtained averaging over the different scores. We obtain, for a particular setting of the parameters and for algorithm ${\cal A}$
$$
\it{ RankScore}({\cal A})=\frac{MAP({\cal A})+AUC({\cal A})+ NDCG({\cal A})+F_1@10({\cal A})+ NDCG@10({\cal A})}{5}
$$
Similarly
$$
\it{ClassScore}({\cal A})=\frac{F_1({\cal A})+A({\cal A})}{2}.
$$
The $\it{RatingScore}$ is defined similarly, but we have to rescale the measures in the range $[0,1]$  to have a measure homogeneous with the others. 

\begin{figure}
	\centering
	\includegraphics[width=14cm]{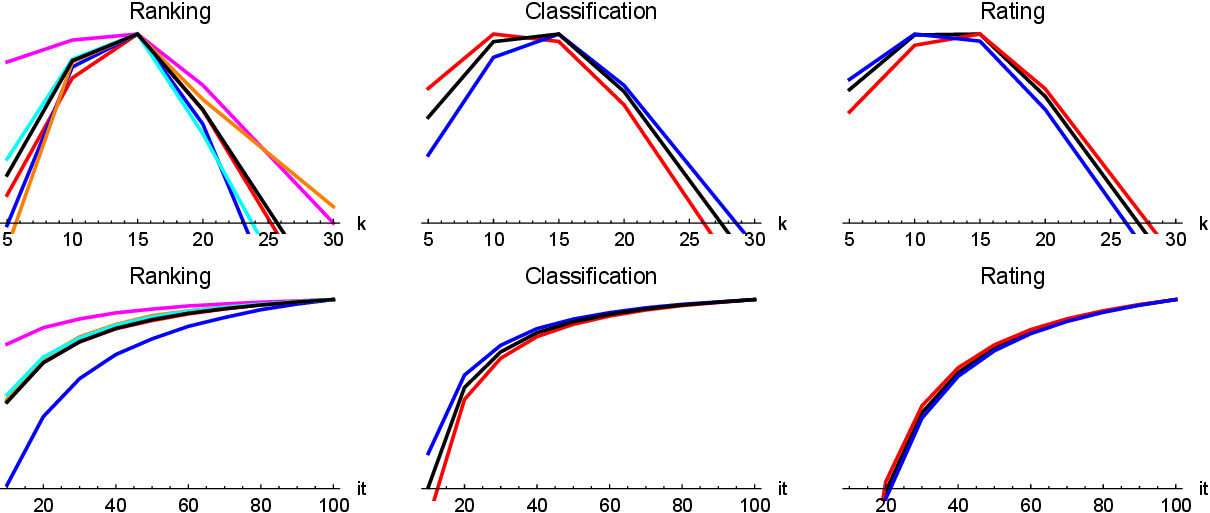}
	\caption{Normalized measures for ranking, classification and rating metrics on the  MovieLens 1M dataset with algorithm PRD. The top plots  show the dependence on $k$,  the bottom plots the behavior of the measures as the number of iterations increases. In each plot, in black are reported the cumulative scores, respectively $\it{RankScore}(PRD)$, $\it{ClassScore}({PRD})$ and $\it{RatingScore}(PRD)$, while the colored lines represent one of the measures in the correspondent class.  We see that all the measures approximately behave similarly achieving the best performance for the same value of $k$, in this case for $k=15$. }\label{figkit}
\end{figure}

To analyze more in detail the correlation  of the different metrics and their dependence on the rank $k$ and on the number of iterations,  we report in Figure~\ref{figkit} (top), for algorithm  PRD, the behavior of the ranking, classification, and rating metrics on the MovieLens 1M dataset for different values of $k$ and the best number of iterations. We have different plots for the  normalized ranking, classification metrics, and rating metrics. Black lines show the trend of the cumulative scores $\it{RankScore}(PRD)$, $\it{ClassScore}({PRD})$ and $\it{RatingScore}(PRD)$. As we can see the different metrics have a very similar behavior and an optimal value of $k=15$ can be clearly identified  independently of the metric adopted. We note, moreover that we can consider the cumulative measures instead that the single ones since they exhibit the same trend. Figure~\ref{figkit} (bottom) shows the trend of the same measures running the algorithm with the best value of $k$, i.e. $k=15$. We plot the different measures as the number of iterations of the ANLS increases. We see that all the scores improve performing more iterations but that with a moderate number of steps the scores are already very good. 
 
 \subsection{Comparison of the different methods}\label{sec:comp}

\begin{figure}
	\centering
	\includegraphics[width=14cm]{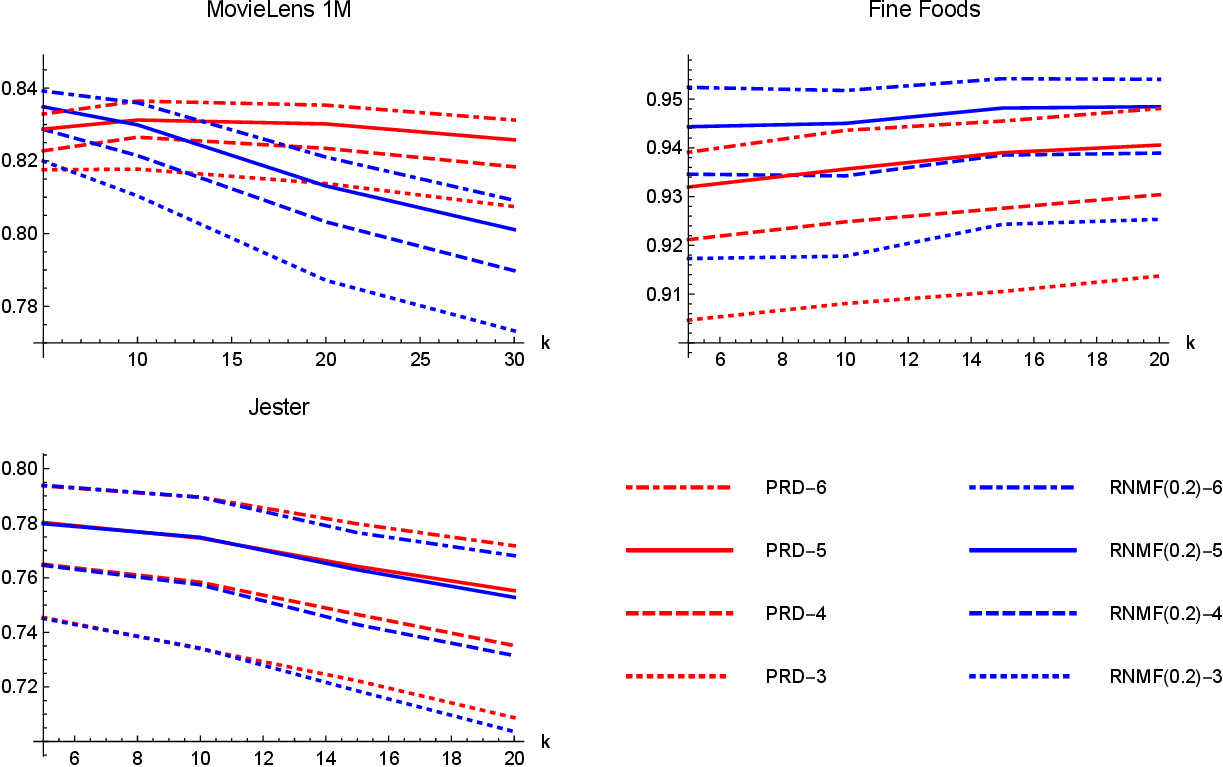}
	\caption{For  the two algorithms PRD and RNMF(0.2) we plot the {\em RankScore} values versus $k$, partitioning the data into training and test set of different sizes.  We consider the $t$-fold cross validation with $t$ ranging from 3 to 6. We see that the trend of the RankScore in relation to the value of $k$ is the same for the different partitioning of the data into training and test set.  }\label{fig-t-cross}
\end{figure}
In this section we analyze in details the performance of the different algorithms proposed in Section~\ref{sec:alg}. We preliminary asses whether  the choice of the parameter $t=5$ in the $t$-cross validation methodology is appropriate.  Figure~\ref{fig-t-cross} shows the trend of the RankScore metric as a function of $k$ for two representative algorithms using values of $t$ ranging from 3 to 6. A value  $t=3$ means that we are using approximately 66\% of the ratings for the training set and a value $t=6$ means that the percentage of ratings in the training set is roughly  $83\%$. From the plots on three datasets we see that there is not significant change in the relative performance of the methods and the optimal value of $k$ is independent of $t$. For this reason, in the following experiments we always use $t=5$. 

We perform a first comparison of the different methods using the cumulative {\it RankScore} metrics to discard some clearly inferior algorithms; then the best version of the remaining algorithms are compared on all proposed measures.

\begin{figure}[!h]
	\centering
	\includegraphics[width=13cm]{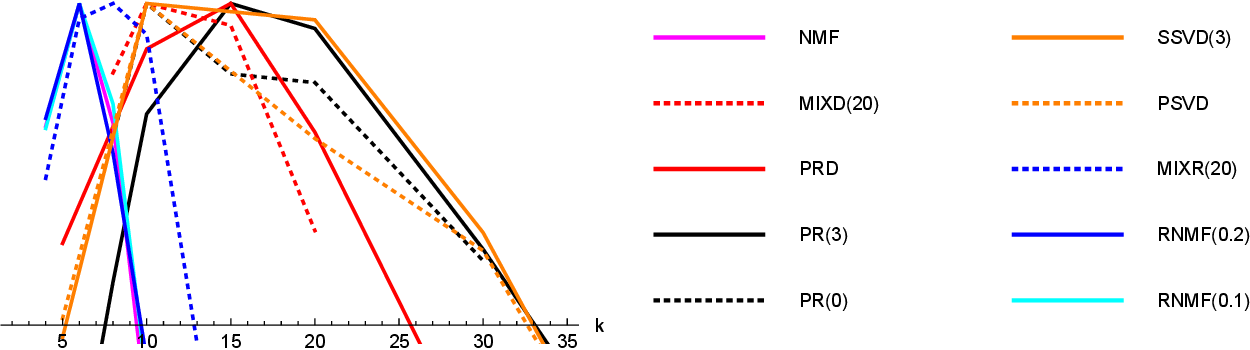}
	\caption{For the optimal value of iterations we show the trend of the normalized {\em RankScore}, plotted  for the MovieLens 1M dataset. Each  algorithm is represented by a specific color.}\label{figfk1}
\end{figure}

In Figure~\ref{figfk1}, for the MovieLens 1M dataset, we report the trend of the normalized {\it RankScore} with respect to $k$. We note that different  methods may have a different optimal~$k$, in particular for
 {RNMF} with $\lambda=0.2$ it is sufficient to take a value of $k$ as small as 5, while for other methods larger values, but never larger than 22, give better results. Although the value of $k$ should denote the number of latent factors in the data, and hence be independent on the algorithm, we note that $k$ contributes to the regularization of the problem. In the light of this observation, it is not surprising that the optimal $k$ is lower for methods with an explicit regularization term, namely RNMF$(\lambda)$ and MIXR.

Some authors~\cite{CKT10} pursuing the SVD approach on the MovieLens 1M dataset, suggest to use $k$ as large as 150. Our tests however suggest that a value $k=10$ suffices for both for PSVD and SSVD(3) methods. The optimal value of $k$ for the {\it RankScore} measure is moderate also for the MovieLens 10M dataset. Note that the use of a smaller $k$ is an obvious computational advantage since the cost of all the algorithms grows linearly with $k$. We investigated whether it is possible to predict for each algorithm a ``good'' value of $k$ given the density of the rating matrix. To this end we summarize in Table~\ref{tab-k} the optimal values of $k$ for different algorithms and datasets. However we could not discern any clear indication based on the density of the matrix and we can only observe that regularized methods works well also with a small $k$. 

\begin{table}[!h]
	\begin{center}
		\begin{tabular}{|l|l|l|l|l|l|l|l|}
			\hline
Dataset& Density& SSVD($\alpha_m$ ) &PR($\alpha_m$) &PRD&RNMF(0.2)&MIXD(20)&MIXR(20)\\ \hline
FineFoods&0.04\%& 40&20&20&2&20&2\\
Movielens 10M&0.21\%&10&10&20&8&20&10\\
Movielens 1M& 4.19\%& 10 &15&15&6&10&8\\
Jester& 53.34\%&10&10&5&5&5&5\\
	\hline
				\end{tabular}
\caption{Optimal values of $k$ for  different datasets and different algorithms.  \label{tab-k}  }	
\end{center}
\end{table}

\begin{figure}
	\centering
	\includegraphics[width=13cm]{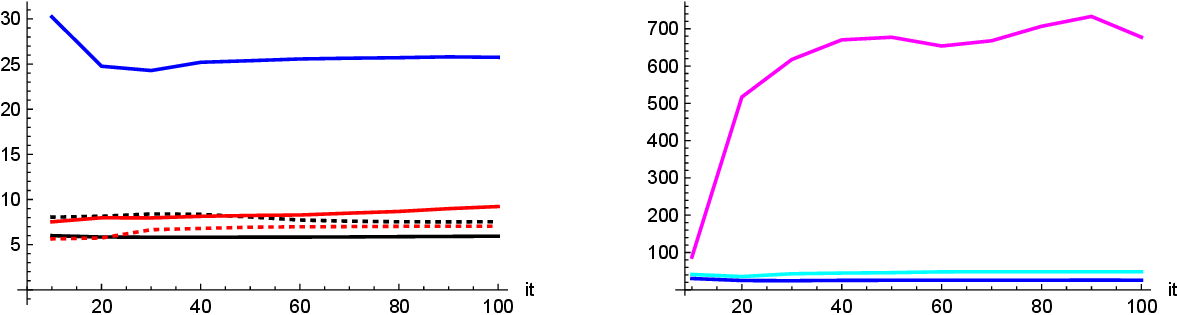}
	\caption{For different algorithms and the optimal value of $k$ we show the maximum value obtained in matrix $B$. In plot (a) using a regularized method (solid blue) or methods with prior. In plot (b) the comparison between NMF (solid pink) --- which is a  method with no explicit regularization or prior term --- and regularized methods such as {RNMF} with $\lambda=0.1$ and $\lambda=0.2$.}\label{figmax}
\end{figure}

Another interesting observation is that methods without explicit regularization work well only if a prior term is used. In Figure~\ref{figmax} we plot the maximum value obtained in the reconstructed recommendation matrix $B$. 
The plot on the left of Figure~\ref{figmax} shows that algorithms incorporating prior terms such as  PR(0), PR(3), PRD  and MIXD(20) are  stable since the maximum is never higher than 10. When using a mild regularization parameter $\lambda=0.2$ we still have  acceptable values of the maximum, even if it tends to be around 25.  In the plot on the right  we see that if no regularization or prior are used, as in the NMF method, the maximum increases to unacceptable levels. Using the regularized method {RNMF} with a tiny value of $\lambda$, such as $\lambda=0.1$,  we get a very good control of the growth of the entries of $B$. For this reason we can rule out, as observed in the literature~\cite{HD11} the simple NMF algorithm used without a prior or a regularization term. 

\begin{center}
 \begin{figure}
 	\centering
 	\includegraphics[width=13cm]{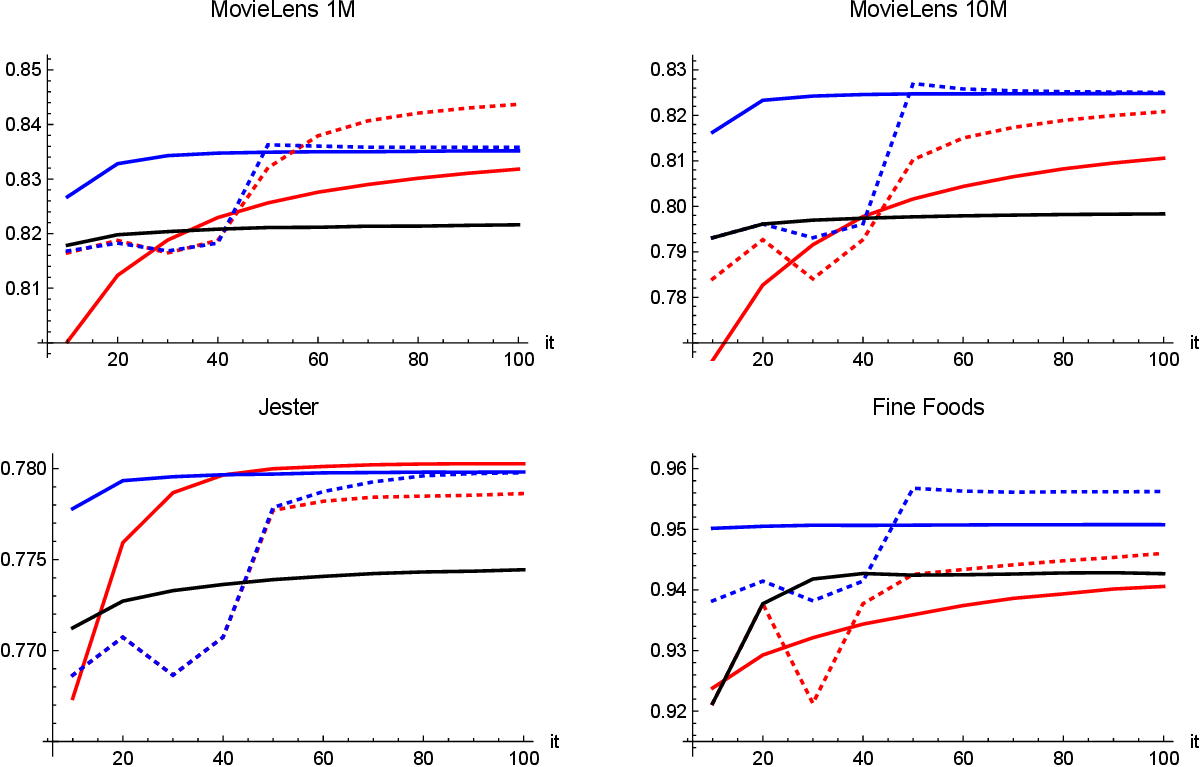}
 	\caption{For the different dataset we show the trend of RankScore as the number of iterations increases. Each algorithm is tested using its own best value of $k$, which is reported in Table~\ref{tab-k}. }\label{figc1}
 \end{figure}
\end{center}

In Figure~\ref{figc1} the direct comparison of the methods respect to  the  cumulative measure {\it RankScore} is plotted for all four datasets. The plots of this figure are relative to the best value of $k$ for that method. These values are summarized in Table~\ref{tab-k} and reported also in Tables~\ref{tab:1m}, \ref{tab:10m}, \ref{tab:j}, \ref{tab:ff}. We see that mixed methods  are in general  superior respect to the corresponding method, except for the Jester dataset where the PRD outperforms MIXD(20) and RNMF(0.2) is better than MIXR(20). We note that mixed techniques perform very well with a moderate value of $k$, and that in general  {\it RankScore}  increases with the number of iterations. For mixed methods such as MIXD(20) and MIXR(20) we see a rapid growth after the first 30-40 iterations, this is implicit in the nature of mixed methods.   In fact, we first use a fast but less accurate method and then we refine adaptively  the provided solution with a more accurate one. These initial steps help in achieving a better performance. In fact, if we compare MIXD(20)  with PRD (with coincides with MIXD(0)) or MIXR(20) with {RNMF}(0.2) (corresponding to MIXR(0)), we clearly see a better performance of those mixed strategies. 
\begin{center}
\begin{table}[!h]
			\begin{tabular}{|l|ll|ll|lllll|}
				\hline
				${\cal A}$,$\quad k$,\quad it&{\em RMSE}&{\em MAE}& {\em F$_1$} &{\em A}& {\em MAP} & {\em AUC}& {\em NDCG} & {\em F$_1@10$} & {\em NDCG}$@10$  \\ \hline		
				RANDOM,-,-&2.181&1.765&0.483&0.487&0.685&0.527&0.827&0.667&0.663\\ \hline
				PSVD,10,-&2.77&2.508&0.067&0.443&0.786&0.681&0.888&0.791&0.785\\
				SSVD($\alpha_m$),10,-&1.077&0.881&0.351&0.534&0.820&0.734&0.905&0.827&0.817\\ \hline				
				PR(0),10,10&{2.825}&{2.567}&0.059&0.440&0.781&0.674&0.886&0.785&0.780\\
				PR($\alpha_m$),15,100&1.072&0.876&0.365&0.540&0.821&0.735&0.905&0.829&0.818\\
				PRD,15,100&0.926&0.715&0.762&0.731&0.832&0.753&0.909&0.839&0.827\\
				{RNMF}(0.2),6,100&0.871&0.680&\cellcolor{lightgray}{0.779}&0.742&0.834&0.761&0.910&0.843&0.829\\ \hline
				MIXD(20),10,100&\cellcolor{lightgray}{0.861}&\cellcolor{lightgray}{0.677}&0.775&\cellcolor{lightgray}{0.743}&\cellcolor{lightgray}{0.843}&\cellcolor{lightgray}{0.743}&\cellcolor{lightgray}{0.916}&\cellcolor{lightgray}{0.850}&\cellcolor{lightgray}{0.839}\\
				MIXR(20),8,50&0.872&0.679&\cellcolor{lightgray}{0.779}&\cellcolor{lightgray}{0.743}&0.835&0.761&0.910&0.844&0.830\\
				\hline
			\end{tabular}
			\caption{Comparisons of the different algorithms on the MovieLens 1M dataset. In gray are highlighted the best performances respect to the error measures. We see that MIXD(20) has the best performance for every measure except for the $F_1$ score where achieves the second best performance. Also the performance of MIXR(20) is always very good. The results obtained using a random matrix are shown for comparison. \label{tab:1m}
}	
	\end{table}
		\end{center}

\begin{table}[!h]
		\begin{center}
				\begin{tabular}{|l|ll|ll|lllll|}
					\hline
	${\cal A}$,$\quad k$,\quad $it$&{\em RMSE}&{\em MAE}& {\em F$_1$} &{\em A}& {\em MAP} & {\em AUC}& {\em NDCG} & {\em F$_1@10$} & {\em NDCG}@10  \\ \hline			
RANDOM,-,-&2.112&1.715&0.457&0.499&0.64&0.524&0.825&0.628&0.689\\ \hline
SSVD($\alpha_m$),10,-&1.005&0.791&0.261&0.560&0.779&0.720&0.897&0.777&0.820\\ \hline
PR($\alpha_m$), 10,100&1.008&0.793&0.256&0.559&0.779&0.720&0.897&0.777&0.820\\
PRD, 20,100&0.919&0.697&0.712&0.722&0.791&0.735&0.902&0.791&0.831\\
RNMF(0.2), 8,100&\cellcolor{lightgray}{0.814}&0.631&0.741&0.742&0.805&\cellcolor{lightgray}{0.762}&\cellcolor{lightgray}{0.908}&0.808&0.841\\ \hline
MIXD(20), 20,100&{0.828}&0.646&0.717&0.728&0.803&0.757&\cellcolor{lightgray}{0.908}&0.803&0.840\\
MIXR(20), 10,50&\cellcolor{lightgray}{0.814}&\cellcolor{lightgray}{0.630}&\cellcolor{lightgray}{0.742}&\cellcolor{lightgray}{0.743}&\cellcolor{lightgray}{0.806}&\cellcolor{lightgray}{0.762}&\cellcolor{lightgray}{0.908}&\cellcolor{lightgray}{0.810}&\cellcolor{lightgray}{0.842}\\ \hline
				\end{tabular}
				\caption{Comparisons of the different algorithms on the MovieLens 10M dataset. In gray are highlighted the best performances respect to the error measures. We see that MIXR(20) has the best performance for every measure, sometime achieving the same score than MIXD(20) or RNMF(0.2). The results obtained using a random matrix are shown for comparison.\label{tab:10m}  }	
	\end{center}
		\end{table}

\begin{table}[!h]
	\begin{center}
		\begin{tabular}{|l|ll|ll|lllll|}
			\hline
			${\cal A}$,$\quad k$,\quad it&{\em RMSE}&{\em MAE}& {\em F$_1$} &{\em A}& {\em MAP} & {\em AUC}& {\em NDCG} & {\em F$_1@10$} & {\em NDCG}@10  \\ \hline	
RANDOM - -&2.111&1.715&0.419&0.513&0.586&0.521&0.82&0.604&0.746\\ \hline
PSVD 5 -&1.487&1.197&0.136&0.605&0.67&0.632&0.861&0.675&0.81\\
SSVD($\alpha_m$ ) 10 -&0.937&0.745&0.544&0.706&0.718&0.712&0.883&0.717&0.841\\
\hline
PR(0) 3 10&1.473&1.189&0.125&0.603&0.674&0.637&0.863&0.677&0.812\\
PR($\alpha_m$) 10 100&0.937&0.745&0.541&0.705&0.719&0.712&0.883&0.717&0.841\\
PRD 5 90&\cellcolor{lightgray}{0.919}&\cellcolor{lightgray}{0.708}&\cellcolor{lightgray}{0.649}&\cellcolor{lightgray}{0.734}&\cellcolor{lightgray}{0.725}&\cellcolor{lightgray}{0.723}&\cellcolor{lightgray}{0.886}&\cellcolor{lightgray}{0.723}&\cellcolor{lightgray}{0.845}\\
RNMF(0.2) 5 100&0.921&\cellcolor{lightgray}{0.708}&\cellcolor{lightgray}{0.649}&\cellcolor{lightgray}{0.734}&0.724&0.722&0.885&\cellcolor{lightgray}{0.723}&\cellcolor{lightgray}{0.845}\\
\hline
MIXD(20) 5 100&0.921&0.709&0.648&0.733&0.723&0.720&0.885&0.722&0.844\\
MIXR(20) 5 100&0.921&\cellcolor{lightgray}{0.708}&\cellcolor{lightgray}{0.649}&\cellcolor{lightgray}{0.734}&0.724&0.722&\cellcolor{lightgray}{0.886}&\cellcolor{lightgray}{0.723}&\cellcolor{lightgray}{0.845}\\
\hline
				\end{tabular}
\caption{Comparisons of the different algorithms on the Jester dataset. In gray are highlighted the best performances respect to the error measures. We see that PRD has the best performance for every measure, sometime achieving the same score than MIXR(20) or RNMF(0.2). The results obtained using a random matrix are shown for comparison.\label{tab:j}  }	
\end{center}
\end{table}

\begin{table}[!h]
	\begin{center}
		\begin{tabular}{|l|ll|ll|lllll|}
			\hline
			${\cal A}$,$\quad k$,\quad it&{\em RMSE}&{\em MAE}& {\em F$_1$} &{\em A}& {\em MAP} & {\em AUC}& {\em NDCG} & {\em F$_1@10$} & {\em NDCG}@10  \\ \hline	
RANDOM - -&2.491&2.007&0.544&0.453&0.904&0.782&0.935&0.88&0.916\\ \hline
PSVD 40 -&3.872&3.500&0.185&0.297&0.937&0.864&0.952&0.937&0.944\\
SSVD($\alpha_m$) 40 -&1.545&1.359&0.377&0.399&0.949&0.891&0.96&0.953&0.954\\
\hline
PR(0) 20 20&4.049&3.734&0.118&0.266&0.936&0.863&0.952&0.935&0.942\\
PR($\alpha_m$) 20 90&1.618&1.459&0.257&0.333&0.951&0.897&0.962&0.950&0.955\\
PRD 20 100&2.715&1.831&0.684&0.620&0.947&0.885&0.961&0.953&0.956\\
RNMF(0.2) 2 90&1.298&0.619&0.906&0.856&0.956&0.905&0.968&0.961&0.964\\
\hline
MIXD(20) 20 100&1.333&1.055&0.553&0.513&0.953&0.902&0.964&0.957&0.959\\
MIXR(20) 2 50&\cellcolor{lightgray}{0.935}&\cellcolor{lightgray}{0.469}&\cellcolor{lightgray}{0.930}&\cellcolor{lightgray}{0.890}&\cellcolor{lightgray}{0.960}&\cellcolor{lightgray}{0.915}&\cellcolor{lightgray}{0.972}&\cellcolor{lightgray}{0.965}&\cellcolor{lightgray}{0.969}\\			\hline
		\end{tabular}
		\caption{Comparisons of the different algorithms on the Fine Foods dataset. In gray are highlighted the best performances respect to the error measures. We see that MIXR(20) has the best performance for every measure. The results obtained using a random matrix are shown for comparison.\label{tab:ff}  }	
	\end{center}
\end{table}

As observed in Section~\ref{subsec:mc} the value of the best $k$ is specific of a method but it is invariant for all the measures considered, while the performance increases with the number of iterations.  Tables~\ref{tab:1m},~\ref{tab:10m}, \ref{tab:j} and \ref{tab:ff} show, for all the interesting algorithms and evaluation metrics, the values achieved for the best value of $k$ and the largest number of iterations~$it$.  As expect mixed methods perform very well, but also {RNMF}(0.2) appears to be a valid choice. Another observation is that there is a big improvement in performance for classification metrics by using more complex methods such as a dynamical prior scheme (PRD) with respect to a static prior scheme such as PR(0) or PR($\alpha_m$). In Table~\ref{tab:10m} we do not report the results for PSVD  and PR(0) which have proved to be less accurate already with the smaller dataset.

The direct comparison of the results provided in this paper with those reported in other papers such as~\cite{CKT10,NK12,SKKR02b} is not possible also for the same datasets. In fact, the testing methodology is in general different and sometimes alternative error measures are  employed. We hope that the use of testing methodologies well established in Information Retrieval, such as the $t$-fold cross validation, will make it easier future comparisons of different methods.

\section{Conclusions} \label{sec:concl}

In this paper we considered the problem of estimating user ratings in Recommender Systems. We proposed new algorithms inspired by classical numerical methods for matrix factorization of the rating matrix, and we compared them with similar approaches in the literature. We considered the quality of the recommendations with respect to ranking, rating and classification metrics, in order to cover all possible usage scenario for a Recommender System. Our first contribution is to show, computing the Spearman correlation between different metrics, that such metrics are usually strongly correlated within the same class, and sometimes even among different classes. 

With extensive experimentations we also observed that, for our datasets, the optimal number of latent factors is moderate (usually below 20), and that the same value is optimal for all the metrics, while, surprisingly enough, this is not an invariant with respect to the method adopted. A possible explanation of this latter phenomenon is that the number of latent factors contributes also to regularize the problem, hence methods with also an explicit regularization work better with a lower number of latent factors. 

Among the proposed new algorithms, mixed strategies, which combine an algorithm with a prior term with a regularized/dynamical method, proved to be the best approaches for tackling the recommendation problem. In particular the MIXR(20) algorithm obtained uniformly good performance for all datasets and evaluation metrics. 

A challenging topic for further research  is to devise mathematical models able to explain some of the behaviors  observed in our experiments and in those of the related literature on the choice of the optimal $k$, and on the global strategy to tackle the recommendation problem. 

\bibliographystyle{plain}

\end{document}